\def\year{2021}\relax
\documentclass[letterpaper]{article} 
\usepackage{aaai21}  
\usepackage{times}  
\usepackage{helvet} 
\usepackage{courier}  
\usepackage[hyphens]{url}  
\usepackage{graphicx} 
\usepackage{graphicx}  
\usepackage{natbib}  
\usepackage{caption} 
\usepackage{booktabs}       
\usepackage{amsfonts}       
\usepackage{nicefrac}       
\usepackage{microtype}      
\usepackage{subfigure}
\usepackage{amsmath}
\usepackage{makecell}
\usepackage{multirow}
\usepackage{amssymb}
\usepackage{varwidth}
\usepackage{caption}
\usepackage[switch]{lineno}

\usepackage{bm}
\usepackage{color}

\allowdisplaybreaks[4]

\newcommand{\tabincell}[2]{\begin{tabular}{@{}#1@{}}#2\end{tabular}}

\title{Interpreting Multivariate Shapley Interactions in DNNs}
\author{
Hao Zhang,\thanks{Contribute equally to this paper.}
Yichen Xie,\footnotemark[1]
Longjie Zheng,
Die Zhang,
Quanshi Zhang\thanks{Corresponding author. This work was done under the supervision of Dr. Quanshi Zhang.}
}
\affiliations{
\\
Shanghai Jiao Tong University, China\\
\{1603023-zh,xieyichen,bugatti,zizhan52,zqs1022\}@sjtu.edu.cn

}

\begin{document}
\maketitle
\begin{abstract}
This paper aims to explain deep neural networks (DNNs) from the perspective of multivariate interactions. In this paper, we define and quantify the significance of interactions among multiple input variables of the DNN. Input variables with strong interactions usually form a coalition and reflect prototype features, which are memorized and used by the DNN for inference. We define the significance of interactions based on the Shapley value, which is designed to assign the attribution value of each input variable to the inference. We have conducted experiments with various DNNs. Experimental results have demonstrated the effectiveness of the proposed method.
\end{abstract}

\section{Introduction}
Deep neural networks (DNNs) have exhibited significant success in many tasks, and the interpretability of DNNs has received increasing attention in recent years.
Most previous studies of post-hoc explanation of DNNs either explain DNN semantically/visually~\cite{lundberg2017unified, ribeiro2016should},
or analyze the representation capacity of DNNs~\cite{higgins2017beta, achille2018emergence, fort2019stiffness, Liang2020Knowledge}.

In this paper, we propose a new perspective to explain a trained DNN, \emph{i.e.} quantifying interactions among input variables that are used by the DNN during the inference process. Each input variable of a DNN usually does not work individually. Instead, input variables may cooperate with other variables to make inferences. We can consider the strongly interacted input variables to form a prototype feature (or a coalition), which is memorized by the DNN.
For example, the face is a prototype feature for person detection, which is comprised of the eyes, nose, and mouth. Each compositional part inside a face does not contribute to the inference individually, but they collaborate with each other to make the combination of all parts a meaningful feature for person detection.

In fact, our research group led by Dr. Q. Zhang have used game-theoretic interactions as new metrics to explain signal-processing behaviors in trained DNNs. Specifically, we have proved that the mathematic connections between interactions and network output~\cite{zhang2020game}, and we have used interactions to explain adversarial samples~\cite{unified2020wang} and the generalization power of DNNs~\cite{zhang2020interpreting}.

Previous studies mainly focused on the interaction between two variables~\cite{lundberg2018consistent, singh2018hierarchical, murdoch2018beyond, janizek2020explaining}.
~\citet{Grabisch1999AnAA} measured $2^n$ different interaction values for all the $2^n$ combinations of $n$ variables, and the cost of computing each interaction value is NP-hard.
In comparison to the unaffordable computational cost, our research summarizes all $2^n$ interactions into a single metric, which provides a global view to explain potential prototype features encoded by DNNs.

In this study, we define the significance of interactions among multiple input variables based on game theory.
In the game theory, each input variable can be viewed as a player. All players (variables) are supposed to obtain a high reward in a game. Let us consider interactions \emph{w.r.t.} the scalar output $y=f(I)$ of the DNN (or interactions \emph{w.r.t.} one dimension of the vectorized network output), given the input $I$ with overall $n$ players. The absolute change of $y$ \emph{w.r.t.} an empty input $\bf 0$, \emph{i.e.} $|f(I)-f(\bf 0)|$, can be used as the total reward of all players (input variables). The reward can be allocated to each player, which indicates the contribution made by the player (input variable), denoted by $\phi_1, \phi_2,\cdots,\phi_n$, satisfies $|f(I)-f({\bf 0})|=\sum_{i=1}^n\phi_i$.

A simple definition of the interaction is given as follows. If a set of $m$ players $S$ always participates in the game together, these players can be regarded to form a coalition. The coalition can obtain a reward, denoted by $\phi_S$. $\phi_S$ is usually different from the sum of rewards when players in the set $S$ participate in the game individually. The additional reward $\phi_S-\sum_{i\in S}\phi_i$ obtained by the coalition can be quantified as the interaction.
If $\phi_S-\sum_{i\in S}\phi_i>0$, we consider players in the coalition have a positive effect. Whereas, $\phi_S-\sum_{i\in S}\phi_i<0$ indicates an negative/adversarial effect among the set of variables in $S$.

However, $\phi_S-\sum_{i\in S}\phi_i$ mainly measures the interaction in a single coalition, whose interaction is either purely positive or purely negative. In real applications, for the set $S$ with $m$ players, players may form at most $2^m-m-1$ different coalitions.
Some coalitions have positive interaction effects, while others have negative interaction effects. Thus, interactions of different coalitions can counteract each other.

\begin{figure*}[t]
\centering
\includegraphics[width=0.8\linewidth, height=0.12\linewidth]{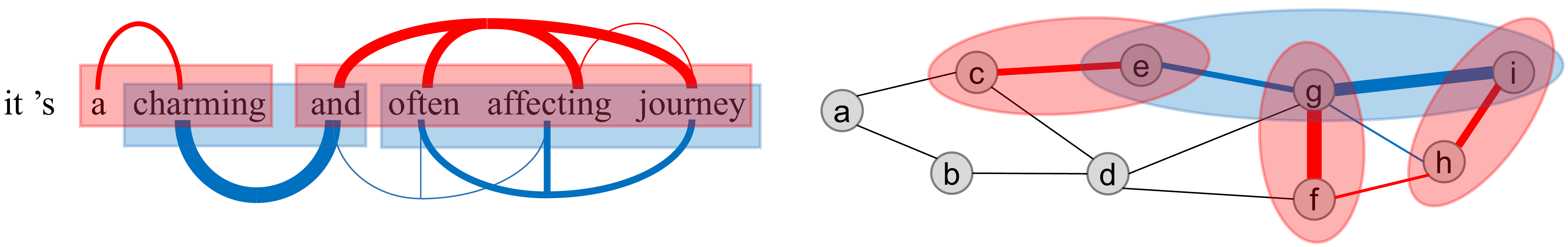}
\caption{Overview of the proposed method. In this paper, we aim to quantify the significance of interactions among a set of input variables. We divide the set of input variables into various coalitions, which mainly encode either positive or negative interaction effects. In this figure, each red region represents a coalition that mainly encodes positive interaction effects, and each blue region represents a coalition that mainly encodes negative interaction effects. Thickness of the edge indicates the strength of the interaction.}
\label{fig:overview}
\end{figure*}

In order to objectively measure the significance of the overall interactions within a specific set $S$ with $m$ players (variables), we propose a new metric, which reflects both positive and negative interaction effects.
We develop a method to divide the $m$ players into a few coalitions, and ensures that each coalition mainly contains positive interaction effects. Similarly, we can also divide the $m$ players into another few coalitions, each of which mainly encodes negative interaction effects.
In this way, we can quantify the significance of both positive and negative interaction effects, as shown in Figure~\ref{fig:overview}.

In experiments, we have applied our method to five DNNs with different architectures for various tasks. Experimental results have provided new insights to understand these DNNs. Besides, our method can be used to mine prototype features towards correct and incorrect predictions of DNNs.

Contributions of this study can be summarized as follows. (1) In this paper, we define and quantify the significance of interactions among multiple input variables in the DNN, which can reflect both positive and negative interaction effects. (2) Our method can extract prototype features, which provides new perspectives to understand the DNN, and mines prototype features towards correct and incorrect predictions of the DNN. (3) We develop a method to efficiently approximate the significance of interactions.

\section{Related work}
\subsection{Interpretability of DNNs in NLP}
The interpretability of DNNs is an emerging direction in machine learning, and many methods have been proposed to explain the DNN in natural language processing (NLP). Some methods computed importance/attribution/saliency values for input variables \emph{w.r.t.} the output of the DNN~\cite{lundberg2017unified, ribeiro2016should, binder2016layerwise, simonyan2013deep, shrikumar2016not, springenberg2014striving, li2015visualizing, li2016understanding}.
Another kind of methods to understand DNNs are to measure the representation capacity of DNNs~\cite{higgins2017beta, achille2018emergence, fort2019stiffness, Liang2020Knowledge, cheng2020explaining}. \citet{guan2019towards} proposed a metric to measure how much information of an input variable was encoded in an intermediate-layer of the DNN.
Some studies designed a network to learn interpretable feature representations.~\citet{chung2017hierarchical} modified the architecture of recurrent neural network (RNN) to capture the latent hierarchical structure in the sentence.~\citet{shen2019ordered} revised the LSTM to learn interpretable representations.
Other studies explained the feature processing encoded in a DNN by embedding hierarchical structures into the DNN~\cite{tai2015improved-0, wang2019tree}.~\citet{wang2019self-attention} used structural position representations to model the latent structure of the input sentence by embedding a dependency tree into the DNN.~\citet{dyer2016recurrent-0} proposed a generative model to model hierarchical relationships among words and phrases.

In comparison, this paper focuses on the significance of interactions among multiple input variables, which provides a new perspective to understand the behavior of DNNs.

\subsection{Interactions}
Several studies explored the interaction between input variables.~\citet{bien2013a} developed an algorithm to learn hierarchical pairwise interactions inside an additive model.~\citet{sorokina2008detecting} proposed a method to detect the statistical interaction via an additive model-based ensemble of regression trees.
Some studies defined different types of interactions between variables to explain DNNs from different perspectives.~\citet{murdoch2018beyond} proposed to use the contextual decomposition to extract the interaction between gates in the LSTM. ~\citet{singh2018hierarchical, jin2019towards} extended the contextual decomposition to produce a hierarchical clustering of input features and contextual independence of input words, respectively.~\citet{janizek2020explaining} quantified the pairwise feature interaction in DNNs by extending the explanation method of Integrated Gradients~\cite{sundararajan2017axiomatic}.
~\citet{tsang2018detecting} measured the pairwise interaction by applying the learned weights of the DNN.

Unlike above studies,~\citet{lundberg2018consistent} defined the interaction between two variables based on the Shapley value for tree ensembles.
Because Shapley value was considered as the unique standard method to estimate contributions of input words to the prediction score with solid theoretical foundations~\cite{weber1988probabilistic}, this definition of the interaction can be regarded to objectively reflect the collaborative/adversarial effects between variables \emph{w.r.t.} the prediction score.
Furthermore,~\citet{Grabisch1999AnAA, dhamdhere2019the} extended this definition to the elementary interaction component and the Shapley-Taylor index among various variables, respectively. These studies quantified the significance of interaction for each possible specific combination of variables, instead of providing a single metric to quantify interactions among all combinations of variables. In comparison, this paper aims to use a single metric to quantify the significance of interactions among all combinations of input variables, and reveal prototype features modeled by the DNN.
\subsection{Shapley values}
The Shapley value was initially proposed in the game theory~\cite{shapley1953value}. Let us consider a game with multiple players. Each player can participate in the game and receive a reward individually. Besides, some players can form a coalition and play together to pursue a higher reward. Different players in a coalition usually contribute differently to the game, thereby being assigned with different proportions of the coalition's reward.
The Shapley value is considered as a unique method that fairly allocates the reward to players with certain desirable properties~\cite{weber1988probabilistic, ancona2019explaining}. Let $N=\{1,2,\cdots,n\}$ denote the set of all players, and $2^{N}$ represents all potential subsets of $N$. A game $v:2^{N}\rightarrow \mathbb{R}$ is implemented as a function that maps from a subset to a real number. When a subset of players $S\subseteq N$ plays the game, the subset can obtain a reward $v(S)$.
Specifically, $v(\emptyset)=0$. The Shapley value of the $i$-th player $\phi_v(i|N)$ can be considered as an unbiased contribution of the $i$-th player.
\begin{small}
\begin{equation}
\label{eqn:shapleyvalue}
\phi_v(i|N)=\sum_{S\subseteq N\setminus\{i\}}\frac{(n-|S|-1)!|S|!}{n!}\Big[v(S\cup\{i\})-v(S)\Big]
\end{equation}
\end{small}~\citet{weber1988probabilistic} have proved that the Shapley value is the only reward with the following axioms.\\
\textbf{Linearity axiom:} If the reward of a game $u$ satisfies $u(S)=v(S)+w(S)$, where $v$ and $w$ are another two games.
Then the Shapley value of each player $i\in N$ in the game $u$ is the sum of Shapley values of the player $i$ in the game $v$ and $w$, \emph{i.e.} $\phi_u(i|N) =\phi_v(i|N)+\phi_w(i|N)$.\\
\textbf{Dummy axiom:} The dummy player is defined as the player that satisfies $\forall S\subseteq N\setminus\{i\}, v(S\cup\{i\})=v(S)+v(\{i\})$. In this way, the dummy player $i$ satisfies $v(\{i\})=\phi_v(i|N)$, \emph{i.e.} the dummy player has no interaction with other players in $N$.\\
\textbf{Symmetry axiom:} If $\forall S\subseteq N\setminus\{i, j\}, v(S\cup\{i\})=v(S\cup\{j\})$, then $\phi_v(i|N)=\phi_v(j|N)$.\\
\textbf{Efficiency axiom:} $\sum_{i\in N}\phi_v(i|N)=v(N)$. The efficiency axiom can ensure the overall reward can be distributed to each player in the game.

\section{Algorithm}
\subsection{Multivariate interactions in game theory}

Given a game with $n$ players $N=\{1,2,\cdots,n\}$, in this paper, we define the interaction among multiple players based on game theory. Specifically, we focus on the significance of interactions of a set of $m$ players selected from all players. Let $2^N$ denote all potential subsets of $N$, \emph{e.g.} if $N=\{a,b\}$, then $2^N=\{\emptyset, \{a\}, \{b\}, \{a,b\}\}$. The game $v:2^N\rightarrow \mathbb{R}$ is a set function that maps a set of players to a scalar value. We consider this scalar value as the reward for the set of players.
For example, we consider the DNN with a scalar output as a game, and consider the input variables as a set of players. In this way, $v(S)$, $S\subseteq N$, represents the output of the DNN when all input variables in $N\setminus S$ are replaced by the baseline value. In this paper, the baseline value is set as zero, which is similar to~\cite{ancona2019explaining}.
If the DNN is trained for multi-category classification, $v(S)$ is implemented as the score of the true category before the softmax layer.
The overall reward can be represented as $v(N)-v(\emptyset)$, \emph{i.e.} the score obtained by all input variables subtract the score obtained by the zero input.
The overall reward can be allocated to each player as $\sum_{i=1}^n\phi_v(i|N)=v(N)-v(\emptyset)$, where $\phi_v(i|N)$ denotes the allocated reward of the player $i$ over the set $N$.
Each player is supposed to obtain a high reward in the game.
In this paper, we use $\phi(i|N)$ to represent $\phi_v(i|N)$ in the following manuscript for simplicity without causing ambiguity.
Specifically, $\phi(i|N)$ can be computed as the Shapley value of $i$ based on Equation~\eqref{eqn:shapleyvalue}. The Shapley value is an unbiased method to allocate the reward, so we use the Shapley value to define the interaction.

\paragraph{Interaction between two players:}
Let us first consider the interaction between two players in the game $v$. Given two players $i$ and $j$, $\phi(i|N)$ and $\phi(j|N)$ represent their Shapley values, respectively. If players $i$ and $j$ always participate in the game together and always be absent together, then they can be regarded to form a coalition. The reward obtained by the coalition is usually different from the sum of rewards when players $i$ and $j$ participate in the game individually.
This coalition can be considered as one singleton player, denoted by $S_{ij}$. Note that the Shapley value of the coalition can also be computed using Equation~\eqref{eqn:shapleyvalue} by replacing the player $i$ with $S_{ij}$.
In this way, we can consider there are only $n-1$ players in the game, including $S_{ij}$, and excluding players $i$ and $j$, $N'=N\setminus\{i,j\}\cup S_{ij}$.
The interaction between players $i$ and $j$ is defined as the additional reward brought by the coalition $S_{ij}$ \emph{w.r.t.} the sum of rewards of each player playing individually, \emph{i.e.}
\begin{small}
\begin{equation}
\label{eqn:TwoInteraction}
B(S_{ij})=\phi(S_{ij}|N')-
\left[\phi(i|N_i)+\phi(j|N_j)\right],
\end{equation}
\end{small}
where {\small$\phi(i|N_i)$}, {\small$\phi(j|N_j)$}, and {\small$\phi(S_{ij}|N')$} are computed over the set of players {\small$N_i=N\setminus\{j\}$}, {\small$N_j=N\setminus\{i\}$}, and {\small$N'$}, respectively.
$\textbf{(1)}$ If {\small$B(S_{ij})>0$}, then players $i$ and $j$ cooperate with each other for a higher reward, \emph{i.e.} the interaction is positive.
$\textbf{(2)}$ If {\small$B(S_{ij})<0$}, then the interaction between players $i$ and $j$ leads to a lower reward, \emph{i.e.} the interaction is negative.
\begin{figure}[t]
\centering
\includegraphics[width=0.99\linewidth]{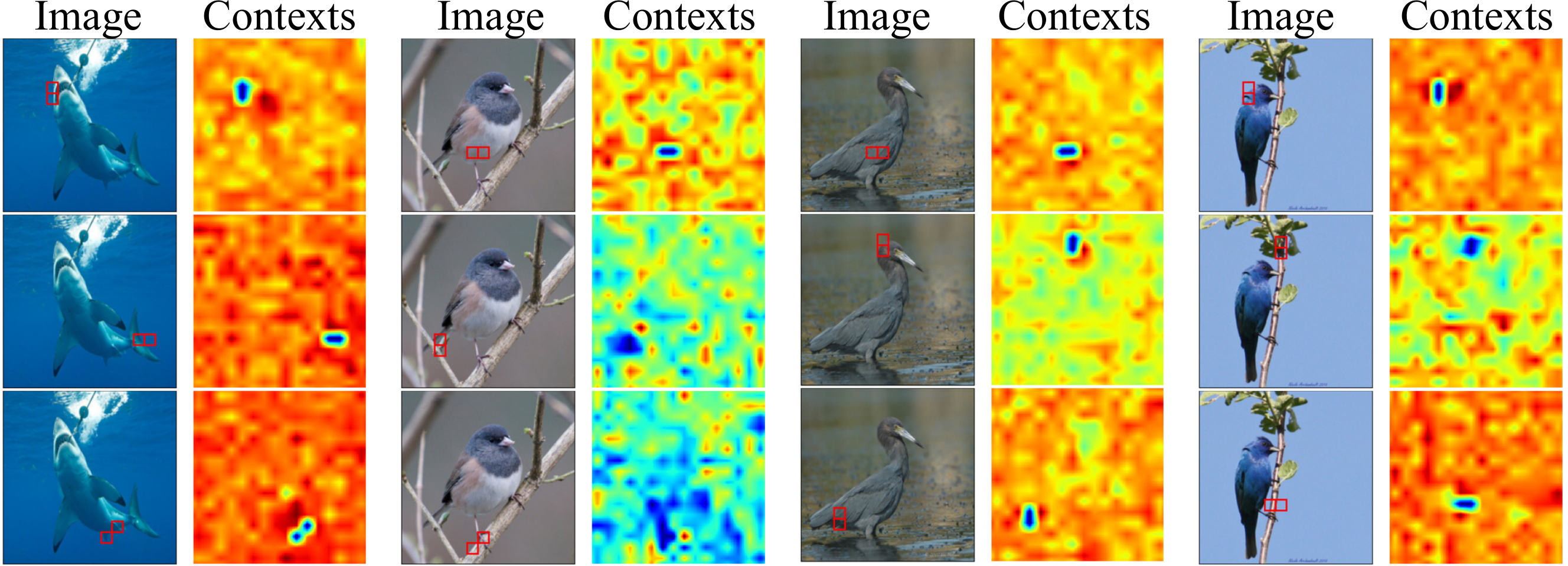}
\caption{Visualization of salient contexts $S$ given different pairs of players.}
\label{fig:contexts}
\end{figure}
{Specifically, we can measure and visualize the interaction between two players in the computer vision (CV) task. We divide the input image into $16\times 16$ grids, and consider each grid as a player. As Figure~\ref{fig:contexts} shows, for a given pair of players $i$ and $j$, we visualized contexts $S$ that boost the strength of the interaction $B(S_{ij})$, \emph{i.e.} $\Delta v(i,j,S)\cdot B(S_{ij})>0$, where $\Delta v(i,j,S)=v(S\cup\{i,j\})-v(S\cup\{i\})-v(S\cup\{j\})+v(S)$.
Let $\text{map}(S)\in \{0,1\}^m$ denote the map corresponding to the context $S$. $\text{map}_k(S)=1$ if the $k$-th grid is in the context $S$; otherwise, $\text{map}_k(S)=0$. We visualize the weighted average contexts as $\sum_S|\Delta v(i,j,S)|\cdot \text{map}(S)$.}

\paragraph{Interaction among multiple players:}
We can extend the definition of interactions to multiple players. Let us focus on a subset of players $A\subsetneqq N$.
If players in $A$ always participate in the game together, then players in $A$ form a coalition, which can be considered as a singleton player, denoted by $[A]$.
The interaction in $A$ can be measured as
\begin{small}
\begin{equation}
\label{eqn:Interaction}
{B([A])=\phi([A]|N_{[A]})-\sum\nolimits_{i\in A}\phi(i|N_i)},
\end{equation}
\end{small}
where $\phi([A]|N_{[A]})$ is the Shapley value of the coalition $[A]$. We compute the Shapley value over the player set $N_{[A]}=N\setminus A\cup\{[A]\}$. Similarly, $\phi(i|N_i)$ represents the Shapley value computed over the set $N_i=N\setminus A\cup\{i\}$, where players in $A\setminus\{i\}$ are supposed not to attend the game.
In this way, $B([A])$ reflects all interactions inside $A$, including both positive and negative interactions, which counteract each other.
For example, let the set $A=\{i,j,k,l\}$, we assume that only the interaction between players $i$ and $j$ and the interaction between players $k$ and $l$ are positive. Interactions of other coalitions are negative. Then, the positive interaction effects are counteracted by negative interaction effects in $B([A])$.
We will introduce more details in Equation~\eqref{eqn:Relationship}.

However, the significance of interactions is supposed to reflect both positive and negative interaction effects.
Therefore, we hope to propose a new metric to measure the significance of interactions, which contains both positive and negative effects among a set of players $A\subsetneqq N$.
We aim to divide all players in $A$ into a number of coalitions $\Omega=\{C_1,C_2,\cdots,C_k\}$, which are ensured to mainly encode positive interaction effects. $\Omega$ is a partition of the set $A$. In the above example, the four players $\{i,j,k,l\}$ is supposed to be divided into $\Omega=\{C_1=\{i,j\},C_2=\{k,l\}\}$.
In this case, the following equation can better reflect positive interaction effects than Equation~\eqref{eqn:Interaction}, when we measure the reward by taking each coalition $C_i$ as a singleton player.
\begin{small}
\begin{equation}
\label{eqn:MaxEffect}
B_{\text{max}}([A])=\max\nolimits_{\Omega}\sum\nolimits_{C\in \Omega}\phi(C|N_C)-\sum\nolimits_{i\in A}\phi(i|N_i)
\end{equation}
\end{small}
where $\phi(C|N_C)$ is computed over $N_{C}=N\setminus A\cup\{[C]\}$, and $\phi(i|N_i)$ is computed over $N_i=N\setminus A\cup\{i\}$.
Similarly, we can use $B_{\text{min}}([A])=\min_{\Omega}\sum_{C\in \Omega}\phi(C|N_C)-\sum_{i\in A}\phi(i|N_i)$ to roughly quantify the negative interaction effects inside the set $A$.
Thus, we define the metric $T([A])$ to measure the significance of both positive and negative interaction effects, as follows.
\begin{small}
\begin{equation}
\begin{aligned}
\label{eqn:Approximate}
T([A])&=B_{\text{max}}([A])-B_{\text{min}}([A])\\
&=\max_{\Omega}\sum_{C\in \Omega}\phi(C|N_C)-\min_{\Omega}\sum_{C\in \Omega}\phi(C|N_C)
\end{aligned}
\end{equation}
\end{small}
\textbf{Relationship between $T([A])$ and $B([A])$:}
Theoretically, the interaction $B([A])$ in Equation~\eqref{eqn:Interaction} can be decomposed into elementary interaction components $I(A)$, which was firstly proposed by~\cite{Grabisch1999AnAA}.
$I(A)$ quantifies the marginal reward of $A$, which removes all marginal rewards from all combinations of $A$.
We derive the following equation to encode the relationship between $B([A])$ and $I(A)$.
\begin{small}
\begin{equation}
\label{eqn:Relationship}
B([A])=\sum\nolimits_{A'\subseteq A, |A'|>1}I(A').
\end{equation}
\end{small}
According to Equation~\eqref{eqn:Relationship}, there are a total of $2^{m}-m-1$ different elementary interaction components inside the interaction $B([A])$, where $m$ represents the number of players inside $A$. Some of these interaction components are positive, and others are negative, which leads to a counteraction among all possible interaction components. In order to quantify the significance of all interaction components, a simple idea is to sum up the absolute values of all elementary interaction components, as follows. $B'([A])=\sum_{A'\subseteq A, |A'|>1}|I(A')|=B^+-B^-$, where $B^+=\sum_{A'\subseteq A,I(A')>0}I(A')$ and $B^-=\sum_{A'\subseteq A,I(A')<0}I(A')$, subject to $|A'|>1$.
In Equation~\eqref{eqn:MaxEffect}, $B_\text{max}([A])$ is supposed to mainly encode most positive interaction effects within $A$, which is highly related to $B^+$. Similarly, $B_\text{min}([A])$ is highly related to $B^-$ for mainly encoding negative interaction effects within $A$. Therefore, $T([A])=B_\text{max}([A])-B_\text{min}([A])$ is highly related to $B'([A])=B^+-B^-$.

\textbf{Why we need a single metric?}~\citet{Grabisch1999AnAA} designed $I(A)$, which computed $2^m$ different values of $I(A')$ for all possible $A'\subseteq A$.
In comparison, the proposed $B([A])$ is a single metric to represent the overall interaction significance among all $m$ input variables, which provides a global view to explain DNNs.

\subsection{Explanation of DNNs using multivariate interactions in NLP}
In this paper, we use the interaction based on game theory to explain DNNs for NLP tasks. Given an input sentence with $n$ words $N=\{1,2,\cdots,n\}$, each word in the sentence can be considered as a player, and the output score of DNN can be taken as the game score $v(S)$. $v(S)$ represents the output of DNN when we mask all words in the set $N\setminus S$, \emph{i.e.} setting word vectors of masked words to zero vectors. For the DNN with a scalar output, $v(S)$ denotes the output of the DNN. If the DNN was trained for the multi-category classification task, we take the output of the true category before the softmax layer as $v(S)$. Strongly interacted words usually cooperate with each other and form a prototype feature, which is memorized by the DNN for inference. Thus, the multivariate interaction can be used to analyze prototype features in DNNs.

Among all input words in $N$, we aim to quantify the significance of interactions among a subset of $m$ sequential words $A \subsetneqq N$.~\citet{chen2018lshapley} showed that non-successive words usually contain much fewer interactions than successive words. Thus, we require each coalition consists of several sequential words to simplify the implementation.
Although $T([A])$ in Equation~\eqref{eqn:Approximate} can be computed by enumerating all possible partitions in $A$, $\Omega$, the computational cost of such a method is too unaffordable.

Therefore, we develop a sampling-based method to efficiently approximate $T([A])$ in Equation~\eqref{eqn:Approximate}.
The computation of $\max_\Omega\sum_{C\in\Omega}\phi(C|N_C)$ requires us to enumerate all potential partitions to determine the maximal value. In order to avoid such computational expensive enumeration, we propose to use $\textbf{p}=\{p_1, p_2, \cdots, p_{m-1}\}$ to represent all possible partitions. $p_i\,(0\le p_i\le 1)$ denotes the probability of the $i$-th word and the $(i+1)$-th word belonging to the same coalition.
We can sample $\textbf{g}=\{g_1, g_2, \cdots g_{m-1}\}$ based on the $\textbf{p}$, \emph{i.e.} $g_i\in\{0, 1\}, \,g_i\sim\text{Bernoulli}(p_i)$, to represent a specific partition $\Omega$. $g_i=1$ indicates the $i$-th word and the $(i+1)$-th word belong to the same coalition.
$g_i=0$ represents these words belong to different coalitions.
In this way, we can use $\text{max}_{\bf p}\mathbb{E}_{\bf g\sim \text{Bernoulli}(\bf p)}\sum_{C\in \Omega_{\bf g}}\phi(C|N_C)$ to approximate $\text{max}_\Omega\sum_{C\in\Omega}\phi(C|N_C)$, where $\Omega_{\bf g}$ denotes the partition determined by $\bf g$.

In addition, $\phi(C|N_C)$ can be approximated using a sampling-based method~\cite{castro2009polynomial}.
\begin{small}
\begin{equation}
\label{eqn:NLPphic}
\begin{aligned}
\phi(C|N_C)&=\underset{r}{\mathbb{E}}\Big[\underset{\substack{S\subseteq N_C,\\|S|=r}}{\mathbb{E}}[v(S\cup C)-v(S)]\Big]\\
&=\underset{r}{\mathbb{E}}\Big\{\underset{\substack{|S|=r+1,\\S\subseteq N_C,S\ni [C]}}{\mathbb{E}}[v(S)]-\underset{\substack{|S|=r,\\S\subseteq N_C,S\not\ni [C]}}{\mathbb{E}}[v(S)]\Big\}
\end{aligned}
\end{equation}
\end{small}
where $r$ represents the number of players fed into the DNN, $N_C=N\setminus A\cup\{[C]\}$.
In this way, we can approximate $\max_\Omega\sum_{C\in\Omega}\phi(C|N_C)$ as $\max_{\bf p}\mathbb{E}_{{\bf g}\sim \text{Bernoulli}({\bf p})}\sum_{C\in\Omega_{\bf g}}\phi(C|N_C)$, as follows.
\begin{small}
\begin{equation}
\begin{aligned}
\label{eqn:NLPphic1}
\mathcal{L}\!\!&=\!\!\!\!\underset{{\bf g}\sim \text{Bernoulli}({\bf p})}{\mathbb{E}}\!\!\sum_{C\in\Omega_{\bf g}}\!\!\!\!\phi(C|N_C),
\max_{\bf p}\mathcal{L}\!=\!
\max_{\bf p}\sum_{i\in A}\\
&\underset{r}{\mathbb{E}}
\Bigg\{\underset{{\bf g}\sim \text{Bernoulli}({\bf p})}{\mathbb{E}}\!\!\!\!\!\!\!\!\lambda_i({\bf g})\Bigg[
\underset{\substack{S\in \text{Sam} (\Omega_\textbf{g}),\\S\ni i, |S|=r+1}}{\mathbb{E}}\!\!\![v(S)]-\!\!\!\!\!\!\underset{\substack{S\in \text{Sam} (\Omega_\textbf{g}),\\S\not\ni i, |S|=r}}{\mathbb{E}}\!\!\![v(S)]
\Bigg]\Bigg\}
\end{aligned}
\end{equation}
\begin{equation}
\label{eqn:gradient}
\begin{aligned}
\frac{\partial \mathcal{L}}{\partial p_i}\!\!=\!\!&\sum_{j\in A}\!\mathbb{E}_r\Big\{\underset{\substack{{\bf g}\sim\text{Bernoulli}({\bf p})\\g_i=1}}{\mathbb{E}}\!\!\!\!\!\!\!\lambda_i({\bf g})\!\Big[\!\!\underset{
\substack{S\in \text{Sam} (\Omega_\textbf{g}),\\S\ni j, |S|=r+1}}{\mathbb{E}}\!\!\!\!\!\!\![v(S)]\!\!-\!\!\!\!\!\underset{\substack{S\in \text{Sam} (\Omega_\textbf{g}),\\S\not\ni j, |S|=r}}{\mathbb{E}}\!\!\!\!\![v(S)]
\Big]\\
-&\!\!\!\!\!\!
\underset{\substack{{\bf g}\sim\text{Bernoulli}({\bf p})\\g_i=0}}{\mathbb{E}}\!\!\!\!\!\!\!\!\lambda_i({\bf g})\Big[\!\!\!\!\underset{
\substack{S\in \text{Sam} (\Omega_\textbf{g}),\\S\ni j, |S|=r+1}}{\mathbb{E}}\!\!\!\!\![v(S)]-\!\!\!\!\!
\underset{\substack{S\in \text{Sam} (\Omega_\textbf{g}),\\S\not\ni j, |S|=r}}{\mathbb{E}}\!\!\!\!\![v(S)]
\Big]\!\!
\Big\}
\end{aligned}
\end{equation}
\end{small}where $\lambda_i({\bf g})=\frac{1}{|C_i|}$, and $S\in \text{Sam}(\Omega_{\bf g})$ represents the sampled set of words, which contains all words in $N\setminus A$ and the coalition $C_i$. $C_i\in\Omega_{\bf g}$ denotes the coalition determined by ${\bf g}$ that contains the word $i$. We learn $\bf p$ to maximize the above equation.
For example, let $A=\{x_1,x_2,x_3,x_4,x_5,x_6\}$, and the sampled ${\bf g}=\{g_1=1, g_2=1, g_3=0, g_4=0, g_5=1\}$. We consider the first three words in $A$ as a coalition, and the last two words in $A$ as another coalition, \emph{i.e.} $\Omega_{\bf g}=\{C_1=\{x_1,x_2,x_3\}, C_2=\{x_4\}, C_3=\{x_5,x_6\}\}$.
The set $S$ is supposed to be sampled over these coalitions in $\Omega_{\bf g}$, instead of over individual words. In this case, we have $\lambda_1({\bf g})=\lambda_2({\bf g})=\lambda_3({\bf g})=1/3$, $\lambda_4({\bf g})=1$, $\lambda_5({\bf g})=\lambda_6({\bf g})=1/2$.
In order to learn $p_i$, we compute $\partial \mathcal{L}/\partial p_i$, which is given in Equation~\eqref{eqn:gradient}.
In this way, we can use $\max_{\bf p}\mathcal{L}$ in Equation~\eqref{eqn:NLPphic1} to approximate the solution to $\max_\Omega\sum_{C\in\Omega}\phi(C|N_C)$. Similarly, we can use $\min_{\bf p}\mathcal{L}$ to approximate the solution to $\min_\Omega\sum_{C\in\Omega}\phi(C|N_C)$, thereby obtaining $T([A])=\max_\Omega\sum_{C\in\Omega}\phi(C|N_C)-\min_\Omega\sum_{C\in\Omega}\phi(C|N_C)$.

\begin{table}[t]
\begin{center}
\caption{Accuracy of the estimated partition.}
\label{tab:toy_example}
\resizebox{0.85\linewidth}{!}{
\begin{tabular}{lccc}
\toprule
& \tabincell{c}{Add-Multiple\\ Dataset} & \tabincell{c}{AND-OR\\ Dataset} & \tabincell{c}{Exponential\\ Dataset} \\
\midrule
Baseline 1 & 0.500 & 0.503 & 0.506 \\
Baseline 2 & 1.000 & 0.996 & 1.000\\
Baseline 3 & 1.000 & 0.523 & 0.846\\
\midrule
Our method & \textbf{1.000} & \textbf{0.999} & \textbf{1.000}\\
\bottomrule
\end{tabular}
}
\end{center}
\end{table}

\textbf{Comparison of computational cost:} Given a set of $m$ words in a sentence with $n$ words, the computation of $T([A])$ based on its definition in Equation~\eqref{eqn:Approximate} and Equation~\eqref{eqn:shapleyvalue} is NP-hard, \emph{i.e.} $O(2^nm)$\footnote[1]{Please see supplementary materials for the proof.}.
Instead, we propose a polynomial method to approximate $T([A])$ with computational cost $O(K_1K_2K_3nm)$\footnotemark[1], where $K_1$ denotes the number of updating the probability $\bf p$, $K_2$ and $K_3$ represent the sampling number of $\bf g\sim\text{Bernoulli}({\bf p})$ and $S\in\text{Sam}(\Omega_{\bf g})$, respectively.
In addition, we conducted experiments to show the accuracy of the estimation of $T([A])$ increases along with the increase of the sampling number.

\section{Experiments}
\textbf{Evaluation of the correctness of the estimated partition of coalitions:}
The core challenge of evaluating the correctness of the partition of coalitions was that we had no ground-truth annotations of inter-word interactions, which were encoded in DNNs. To this end, we constructed three new datasets with ground-truth partitions between input variables, \emph{i.e.} \emph{the Addition-Multiplication Dataset}, \emph{the AND-OR Dataset}, and \emph{the Exponential Dataset}.

\begin{figure}
\begin{center}
\includegraphics[width=0.75\linewidth]{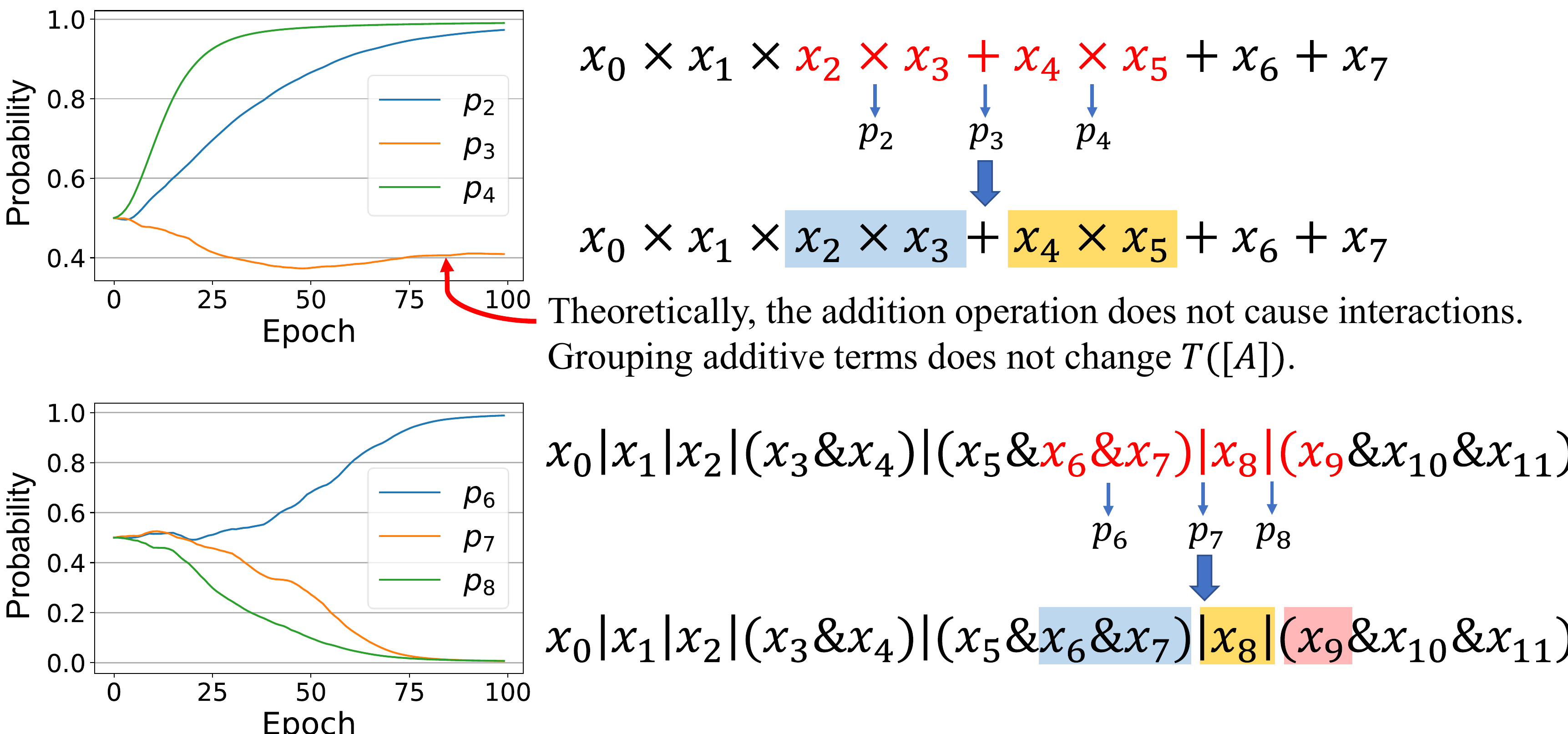}
\captionof{figure}{Convergence of $p_i$ during the training process.}
\label{fig:toy_example}
\end{center}
\end{figure}

$\bullet$ \emph{The Addition-Multiplication Dataset} contained $10000$ addition-multiplication models, each of which only consisted of addition operations and multiplication operations. For example, $y=f({\bf x})=x_1+x_2\times x_3+x_4\times x_5+x_6+x_7$.
Each variable $x_i$ was a binary variable, \emph{i.e.} $x_i\in\{0,1\}$. For each model, we selected a number of sequential input variables to construct $A$, \emph{e.g.} $A=\{x_2,x_3,x_4,x_5\}$. We applied our method to extract the partition of coalitions \emph{w.r.t.} $y$, which maximized $\sum_{C\in\Omega}\phi(C|N_C)$.

\begin{figure*}[!ht]
\centering
\includegraphics[width=0.99\linewidth, height=0.16\linewidth]{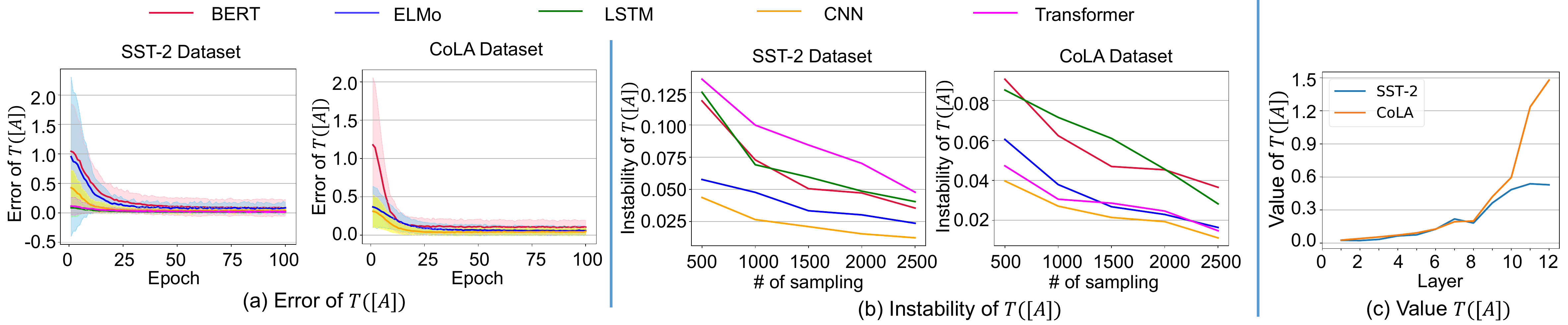}
\caption{(a) The decrease of error of $T([A])$ during the learning process. We obtained accurate $T([A])$ after 100 epochs. (b) The instability of interactions with different numbers of sampling. When the number of sampling was larger than $1000$, the instability of $T([A])$ was less than $0.1$ on all models and datasets.
(c) The significance of interactions in the computation of intermediate-layer features gradually increased during the forward propagation in the BERT.}
\label{fig:instabilityAndCorrectness}
\end{figure*}

$\bullet$ \emph{The AND-OR Dataset} contained $10000$ AND-OR models, each of which only contained AND operations and OR operations. For simplicity, we used $\&$ to denote the AND operation, and used $|$ to denote the OR operation. For example, $y=f({\bf x})=x_1\; |\;( x_2\; \&\; x_3)\;|\;( x_4\; \&\; x_5)\;|\; x_6\; |\; x_7$.
Each variable $x_i$ was a binary variables, \emph{i.e.} $x_i\in\{0,1\}$.
For each model, we selected a number of sequential input variables to construct $A$, \emph{e.g.} $A=\{x_2,x_3,x_4,x_5\}$. We used our method to extract the partition of coalitions, which maximized $\sum_{C\in\Omega}\phi(C|N_C)$.

$\bullet$ \emph{The Exponential Dataset} contained $10000$ models, each of which contained exponential operations and addition operations.  For example, $y=f({\bf x})=x_1^{x_2}+x_3^{x_4}+x_5+x_6$.
Each variable $x_i$ was a binary variables, \emph{i.e.} $x_i\in\{0,1\}$.
For each model, we selected a number of sequential input variables to construct $A$, which always contained the exponential operations, \emph{e.g.} $A=\{x_1,x_2,x_3,x_4\}$. We used our method to extract the partition of coalitions, which maximized $\sum_{C\in\Omega}\phi(C|N_C)$.

We compared the extracted partition of coalitions with the ground-truth partition of coalitions. Variables in multiplication operations and AND operations usually had positive interactions, which were supposed to be allocated into one coalition during $\max_\Omega\sum_{C\in\Omega}\phi(C|N_C)$. Variables in OR operations had negative interactions, which were supposed to be allocated to different coalitions during $\max_\Omega\sum_{C\in\Omega}\phi(C|N_C)$. Note that the addition operation did not cause any interactions. Therefore, we did not consider interactions of addition operations when we evaluated the correctness of the partition.
In the above example of addition-multiplication model, if we considered the set $A=\{x_2,x_3,x_4,x_5\}$, then the ground-truth partition was supposed to be either $\{\{x_2,x_3\},\{x_4,x_5\}\}$ or $\{\{x_2,x_3,x_4,x_5\}\}$.
For the above example of AND-OR model and $A=\{x_2,x_3,x_4,x_5\}$, the ground-truth partition should be $\{\{x_2,x_3\},\{x_4,x_5\}\}$.
For each operation, if the proposed method allocated variables of an operation in the same way as the ground-truth partition, then we considered the operation was correctly allocated by the proposed method. The average rate of correctly allocated operations over all operations was reported in Table~\ref{tab:toy_example}.

\begin{figure*}[!ht]
\centering
\includegraphics[width=0.87\linewidth]{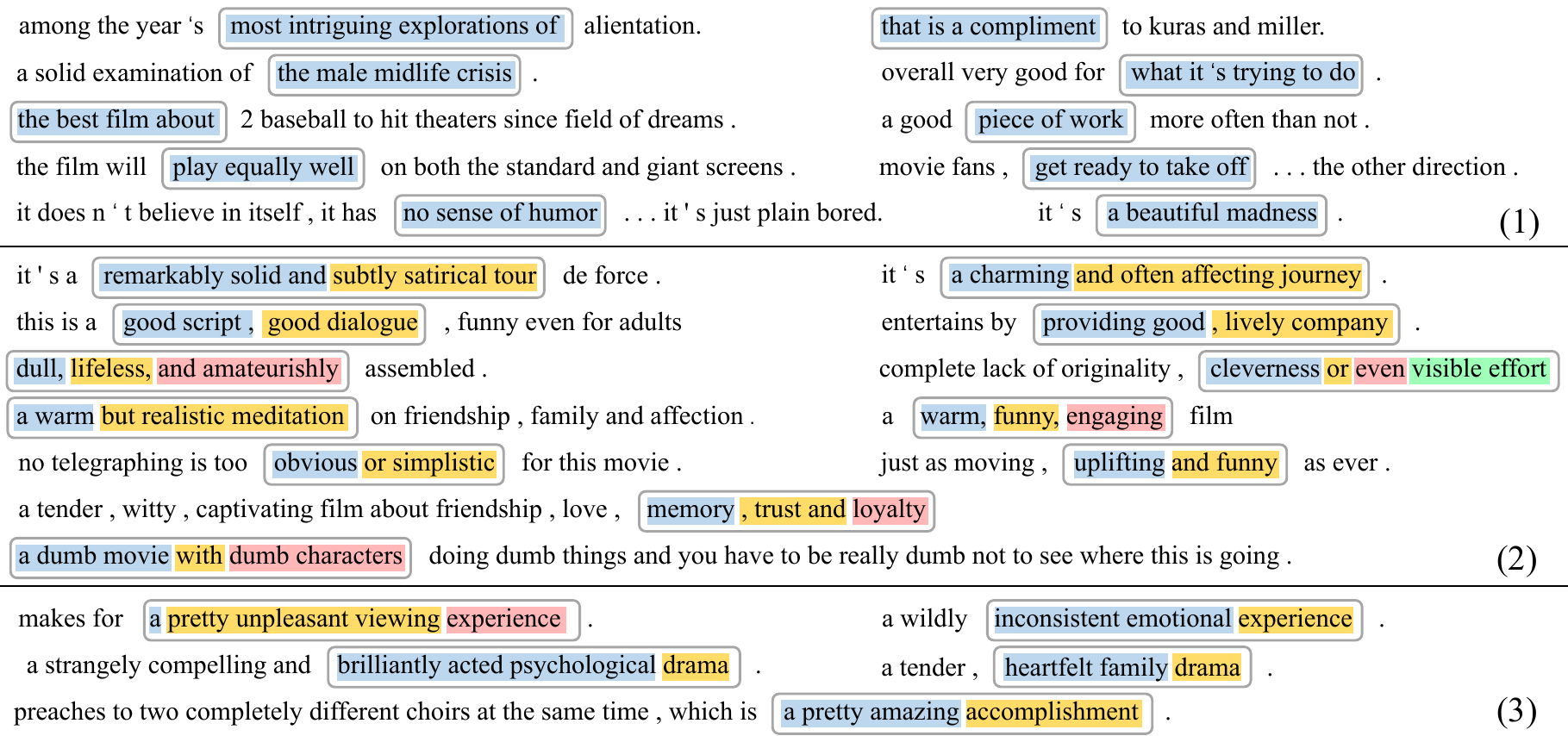}
\caption{Prototype features modeled by the BERT trained using the SST-2 dataset. Grey box indicated words in $A$. We used different colors to indicate the extracted coalitions. Words in the same coalition formed a prototype feature modeled by the DNN. The set of words $A$ was randomly selected.}
\label{fig:interaction_example}
\end{figure*}

We also designed three baselines.
\textbf{Baseline 1} randomly combined input variables to generate coalitions.~\citet{lundberg2018consistent} defined the interaction between two input variables, which was used as the \textbf{Baseline 2}.~\citet{li2016understanding} proposed a method to estimate the importance of input variables, which was taken as \textbf{Baseline 3}. \textbf{Baseline 2} and \textbf{Baseline 3} merged variables whose interactions were greater than zero to generate coalitions.

Table~\ref{tab:toy_example} compares the accuracy of coalitions generated by our method and that of other baselines. Our method achieved a better accuracy than other baselines.
Figure~\ref{fig:toy_example} shows the change of the probability $p_i$ during the training process.
Our method successfully merged variables into a single coalition in multiplication operations and AND operations. Besides, our method did not merge variables in OR operations. Note that the probability $p_3$ of the addition operation in Figure~\ref{fig:toy_example} did not converge. It was because the addition operation did not cause any interactions, \emph{i.e.} arbitrary value of $p_3$ satisfied the ground-truth.

\textbf{Evaluation of the accuracy of $T([A])$:}
We applied our method to DNNs in NLP to quantify the interaction among a set of input words. We trained DNNs for two tasks, \emph{i.e.} binary sentiment classification based on the SST-2 dataset~\cite{socher2013recursive} and prediction of linguistic acceptability based on the CoLA dataset~\cite{warstadt2018neural}.
For each task, we trained five DNNs, including the BERT~\cite{devlin2018bert}, the ELMo~\cite{peters2018deep}, the CNN proposed by~\cite{kim2014convolutional}, the two-layer unidirectional LSTM~\cite{hochreiter1997long}, and the Transformer~\cite{vaswani2017attention}.
For each sentence, the set of successive words $A$ was randomly selected.

We compared the extracted significance of interactions $T([A])$ with the accurate significance of interactions. The accurate significance of interactions was quantified based on the definition in Equation~\eqref{eqn:Approximate} and Equation~\eqref{eqn:shapleyvalue}, which was computed by enumerating all partitions $\Omega$ and all subsets $S$ with an extremely high computational cost. Considering the unaffordable computational cost, such evaluation could only be applied to sentences with less than $12$ words.
Figure~\ref{fig:instabilityAndCorrectness} (a) reports the error of $T([A])$, \emph{i.e.} $|T_\text{truth}([A])-T([A])|$, where $T_\text{truth}([A])$ was the true interaction significance accurately computed via massive enumerations. We found that the estimated significance of interactions was accurate enough after the training of $100$ epochs.

\textbf{Stability of $T([A])$:}
We also measured the stability of $T([A])$, when we computed $T([A])$ multiple times with different sampled sets of $\bf g$ and $S$. The instability of $T([A])$ was computed as
$\text{instability}=\mathbb{E}_{I}[\frac{\mathbb{E}_{u,v:u\neq v}|T_{(u)}([A])-T_{(v)}([A])|}{\mathbb{E}_{w}|T_{(w)}([A])|}]$, where $T_{(u)}([A])$ denotes the $u$-th computation result of $T([A])$.
Figure~\ref{fig:instabilityAndCorrectness} (b) shows the instability of interactions in different DNNs and datasets.
Experimental results showed that interactions in CNN and ELMo converged more quickly. Moreover, instability decreased quickly along with the increase of the sampling numbers. Our metric was stable enough when the number of sampling was larger than $2000$.

\begin{figure*}
  \centering
  \includegraphics[width=0.99\linewidth]{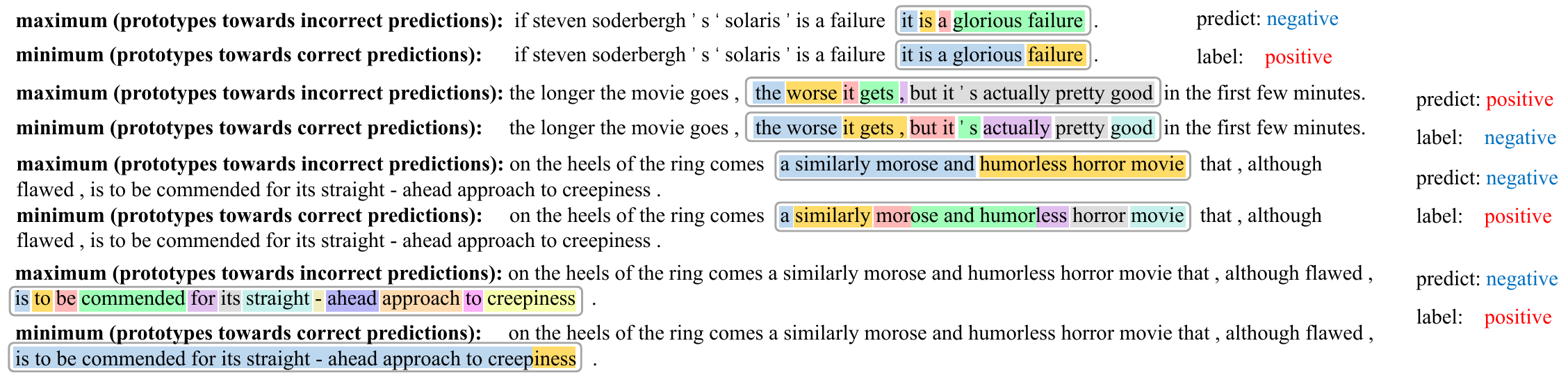}
  \caption{Positive and negative interaction effects of sentences, which were mistakenly classified by the DNN. These results reflect prototype features towards correct and incorrect predictions, and representation flaws of the DNN.}
  \label{fig:fail_example}
\end{figure*}

\textbf{Prototype features modeled by DNNs:}
We analyzed results obtained by the proposed method to explore prototype features modeled by DNNs. Figure~\ref{fig:interaction_example} shows several results on the BERT. We found that: (1) For an entire constituent in the sentence, such as a short clause or a noun phrase, the BERT usually took the whole constituent as a single coalition, \emph{i.e.} a prototype feature. (2) For the set of words that contained conjunctions or punctuations, such as ``and'', ``or'', ``,'', \emph{etc.}, the BERT divided the set at conjunctions or punctuations to generate different coalitions (prototype features). (3) For the constituent modified by multiple adjectives and adverbs, the BERT usually merged all adjectives and adverbs into one coalition, and took the modified constituent as another coalition.
Such phenomena fit the syntactic structure parsing according to human knowledge.

\textbf{Interactions \emph{w.r.t.} the intermediate-layer feature:}
We could also compute the significance of interactions among a set of input words \emph{w.r.t.} the computation of an intermediate-layer feature $\bm f$.
The reward was computed using the intermediate-layer feature. Let ${\bm f}_N$ and ${\bm f}_S$ represent intermediate-layer features obtained by the set of input variables $N$ and $S$, respectively.
Since the intermediate-layer feature usually can be represented as high dimensional vertors, instead of a scalar value. We computed the output $v(S)$ as $v_{\bm f}(S)=\langle{\bm f}_N, {\bm f}_S\rangle/\Vert{\bm f}_N\Vert_2$, where $\Vert{\bm f}_N\Vert_2$ was the L2-norm and was used for normalization.
Figure~\ref{fig:instabilityAndCorrectness} (c) shows the significance of interactions computed with output features of different layers. The significance of interactions among a set of input variables increased along with the layer number. This phenomenon showed that the prototype feature was gradually modeled in deep layers of the DNN.

\textbf{Mining prototype features towards incorrect predictions of DNNs:} The proposed method could be used to analyze sentences, which were mistakenly classified by the DNN. We used the BERT learned using the SST-2 dataset.
Figure~\ref{fig:fail_example} shows two partitions that maximized and minimized the Shapley value of coalitions for each sentence.
For the partition that maximized the Shapley value, the generated coalitions were usually used to explain prototype features toward incorrect predictions of the DNN. In comparison, coalitions generated by minimizing the Shapley value represented prototype features towards correct predictions. For example, we tested two different sets of input variables $A$ for the sentence ``on the heels of ...... to creepiness.''. This sentence was incorrectly predicted to be negative by the DNN. As Figure~\ref{fig:fail_example} shows, the DNN used ``humorless horror movie'' as a prototype features, which led to the incorrect negative prediction. In comparison, if we minimized the Shapley value, we got the coalition ``is to be recommanded for its straight-ahead approach to creep'', which towards the correct positive prediction.

\section{Conclusion}
In this paper, we have defined the multivariate interaction in the DNN based on game theory. We quantify the interaction among multiple input variables, which reflects both positive and negative interaction effects inside these input variables. Furthermore, we have proposed a method to approximate the interaction efficiently. Our method can be applied to various DNNs for different tasks in NLP.
Note that the quantified interaction is just an approximation of the accurate interaction in Equation~\eqref{eqn:Approximate}. Nevertheless, experimental results have verified high accuracy of the approximation.
The proposed method can extract prototype features modeled by the DNN, which provides a new perspective to analyze the DNN.

\section*{Acknowledgements}
This work is partially supported by National Natural Science Foundation of China (61906120 and U19B2043).

\section*{Ethical Impact}
This study has broad impacts on the understanding of signal processing in DNNs. Our work provides researchers in the field of explainable AI with new mathematical tools to analyze DNNs. Currently, existing methods mainly focus on interactions between two input variables. Our research proposes a new metric to quantify interactions among multiple input variables, which sheds new light on the understanding of prototype features in a DNN. We also develop a method efficiently approximate such interactions. As a generic tool to analyze DNNs, we have applied our method to classic DNNs and have obtained several new insights on signal processing encoded in DNNs for NLP tasks.

\bibliography{example_paper}
\end{document}